\documentclass[letterpaper, 10 pt, conference]{ieeeconf}  

\IEEEoverridecommandlockouts                             

\usepackage{cite}
\usepackage{amsmath,amssymb,amsfonts,bm}
\usepackage{algorithmic}
\usepackage{graphicx}
\usepackage{tabularx}
\usepackage{textcomp}
\usepackage{xcolor}
\usepackage[dvipsnames]{xcolor}

\usepackage{booktabs}
\usepackage{multirow}
\usepackage{mathtools}          
\usepackage{mathrsfs}           

\usepackage{graphicx}           
\usepackage{subcaption}         
\usepackage[space]{grffile}     
\usepackage{url}                

\usepackage{blindtext}
\usepackage{wrapfig}
\usepackage{xspace}

\usepackage{float}
\usepackage[capitalize,noabbrev]{cleveref}
\crefname{section}{Sec.}{Secs.}
\Crefname{section}{Section}{Sections}
\Crefname{table}{Table}{Tables}
\crefname{table}{Tab.}{Tabs.}

\newcommand{\infdel}{{\textit{inference delay}}\xspace}

\newcommand{\furuta}{{\textit{\sc Furuta} pendulum}\xspace}
\newcommand{\phihp}{{PHiHP}\xspace} 
 
\newcommand{\rthcp}{{\sc rt-hcp}\xspace}
\newcommand{\tdmpc}{{\sc td-mpc}\xspace}
\newcommand{\rttdmpc}{{\sc rt-tdmpc}\xspace}
 
\newcommand{\pets}{{\sc pets}\xspace}
\newcommand{\rtpets}{{\sc rt-pets}\xspace}
\newcommand{\tddd}{{\sc td3}\xspace}

\newcommand{\dmdp}{delay-MDP\xspace}

\newcommand{\eg}{\textit{e.g.} }
\newcommand{\ie}{\textit{i.e.} }
\begin{document}

\title{\LARGE \bf
RT-HCP: Dealing with Inference Delays and Sample Efficiency to Learn Directly on Robotic Platforms
}

\author{Zakariae El{\ }Asri$^{1}$, Ibrahim Laiche$^{1}$, Clément Rambour$^{1}$, Olivier      Sigaud$^{1}$, Nicolas Thome$^{1,2}$
\thanks{Authors are with $^{1}$Sorbonne Universit\'{e}, CNRS, ISIR, F-75005 Paris, France. Emails: {\tt\small \{elasri, laiche, rambour, sigaud, thome\}@isir.upmc.fr}} %
\thanks{ $^{2}$  Institut Universitaire de France (IUF)  %
}%
}

\maketitle
\thispagestyle{empty}
\pagestyle{empty}

\begin{abstract}
Learning a controller directly on the robot requires extreme sample efficiency. Model-based reinforcement learning (RL) methods are the most sample efficient, but they often suffer from a too long inference time to meet the robot control frequency requirements. In this paper, we address the sample efficiency and inference time challenges with two contributions. First, we define a general framework to deal with inference delays where the slow inference robot controller provides a sequence of actions to feed the control-hungry robotic platform without execution gaps. Then, we compare several RL algorithms in the light of this framework and propose RT-HCP, an algorithm that offers an excellent trade-off between performance, sample efficiency and inference time. We validate the superiority of RT-HCP with experiments where we learn a controller directly on a simple but high frequency FURUTA pendulum platform. Code: github.com/elasriz/RTHCP
\end{abstract}
\IEEEpeerreviewmaketitle

\section{Introduction}

Reinforcement Learning (RL) is a powerful framework for autonomous decision-making, enabling significant success in various domains from video games \cite{Ye2021MasteringAG} to robotics control \cite{Duan2016BenchmarkingDR}.
However, despite significant progress in simulated environments, applying RL algorithms to real-world robotic systems remains highly challenging due to practical constraints such as sample inefficiency, high computational demands, and inference delays \cite{dulac2019challenges}.
To train an RL agent directly on a physical robot, it must learn from limited data while ensuring real-time inference at the control frequency of the system.
In simulated environments, these constraints are often overlooked since agents can generate unlimited data and the environment can wait for computations to finish. In contrast, real-world robotic applications impose strict time and data constraints, making direct learning on hardware significantly more difficult.

This work aims to enable RL deployment on real robots systems while addressing two fundamental constraints: (i) a limited budget on training data, as real-world data collection is costly and time-consuming, and (ii) real-time execution constraints, where the limited processing power of embedded hardware introduces inference delays that must be managed to maintain synchronization with the system’s operating frequency.

\begin{figure}[ht]
\centering
    \begin{subfigure}{0.8\linewidth}
    \centering
        \includegraphics[width=\linewidth]{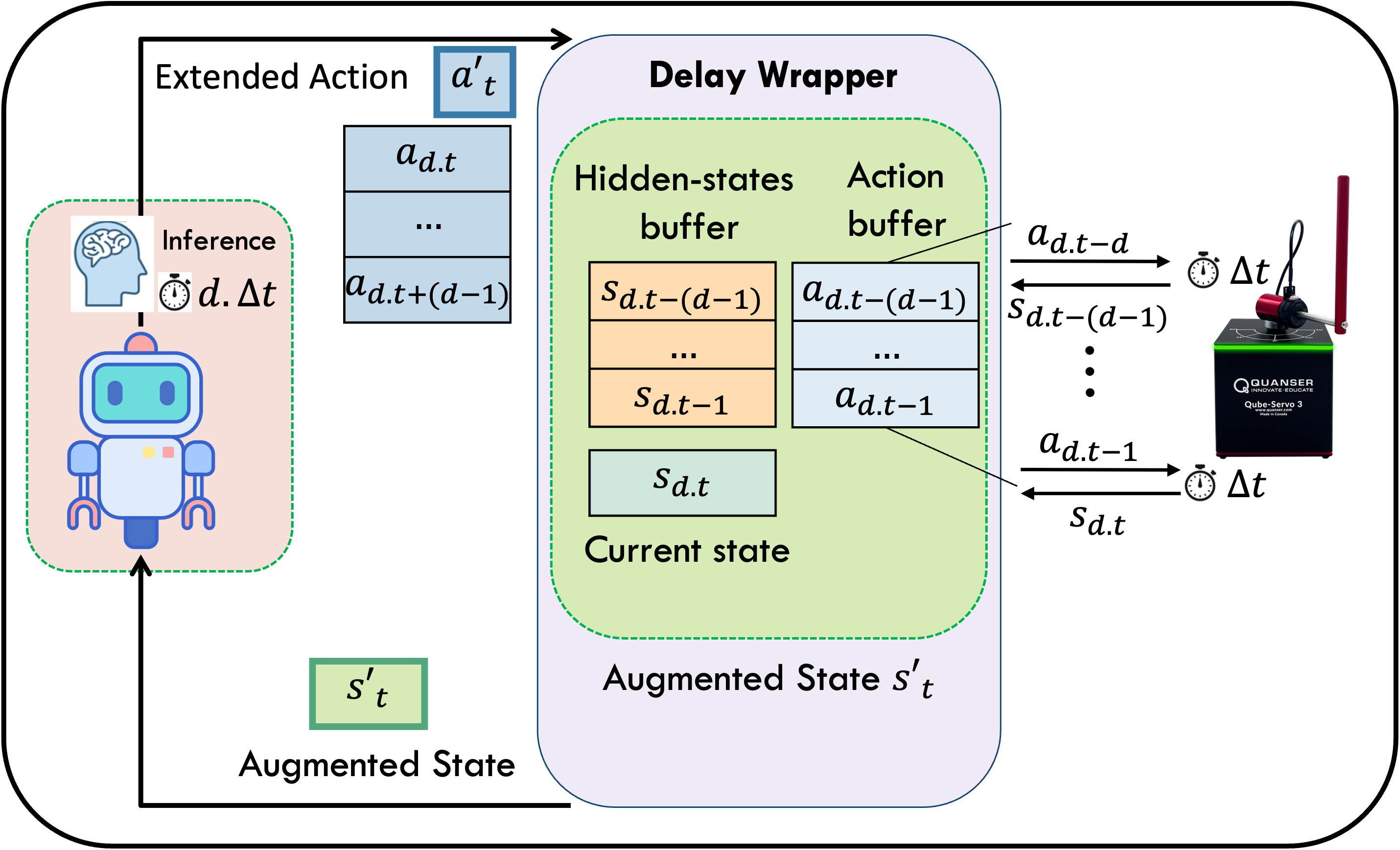}
        \centering
            \caption{Collect trajectories with d-step MPC}
        \label{figure:dMPC}
    \end{subfigure}
    \begin{subfigure}{\linewidth}
        \centering
        \includegraphics[width=0.9\linewidth]{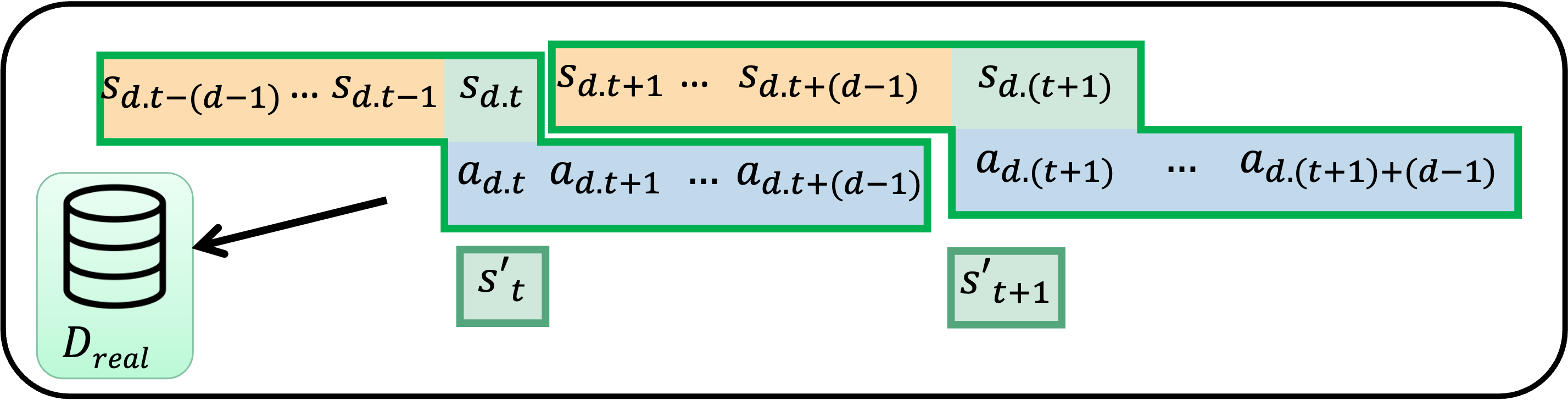}
        \centering\caption{Concatenate augmented states to restore the original MDP}
        \label{figure:dMDP}
    \end{subfigure}
\caption{In the case of inference delay, the agent requires $d$ timesteps to compute the next action, causing it to miss $d-1$ environment steps. The {\bf d-step MPC} framework allows the agent to compensate for this delay by planning a sequence of $d$ actions instead of a single action, and to restore the Markov property by augmenting the current state with the $d$ missed states and the $d$ future actions.}
\label{figure:main_v2}
\end{figure}

Model-free RL (MFRL) methods show strong performance in simulated robotic tasks \cite{ Duan2016BenchmarkingDR}, but they are notoriously sample-inefficient, requiring millions of interactions to learn effective policies. This makes real-world deployment impractical, as data collection on physical robots is always time consuming, often expensive, and even sometimes risky \cite{dulac2019challenges}.

An alternative approach, model-based RL (MBRL) \cite{chua2018deep
}, has gained attention for its ability to learn system dynamics from limited data and use predictive models for planning. Unlike model-free methods, MBRL builds an internal representation of the environment, allowing the agent to simulate potential future states and make more informed decisions. A common strategy in MBRL is Model Predictive Control (MPC), which optimizes control actions over a finite horizon by leveraging a learned forward model to predict future states \cite{pets2019imp, Hansen2022tdmpc}. Since MPC requires solving an optimization problem at each decision step, many approaches rely on the Cross-Entropy Method (CEM) \cite{Boer2005ATO}, a gradient-free stochastic optimization technique, to efficiently search for optimal sequences of actions, particularly in complex and nonlinear dynamics.

While MBRL significantly improves sample efficiency, it comes at the cost of high computational demands, often requiring considerable inference time to compute actions.
As a result, the agent may need multiple control timesteps to compute an action, misaligning its decision-making cycle with the system’s control loop. This results in {\em execution gaps} where no new control commands are issued, potentially degrading performance and stability. We refer to this problem as \infdel, a fundamental challenge in deploying MBRL for real-time control.

It has recently been shown that inference delays could be drastically reduced by combining MPC with model-free RL \cite{Hansen2022tdmpc} and that an even better compromise between inference time, sample efficiency, and performance could be found by also leveraging prior knowledge about the dynamics of the controlled system \cite{asri2024physics}.
However, while these approaches reduce inference delays, they do not eliminate them entirely. A large inference time persists, especially in real-world scenarios in which embedded computational resources are limited, making \infdel a persistent challenge for real-world robotic applications where real-time control is critical.

Unlike transmission delays, which simply create a lag between action selection and execution while maintaining the control frequency of the system, \infdel arises from the agent’s decision-making process, disrupting synchronization between its decision-making cycle and the control frequency of the system. 

To address the challenge of \infdel in real-world robotics, we present the following key contributions:

\begin{enumerate}
    \item We introduce a \dmdp problem to formally account for \infdel and propose a d-step MPC framework for MBRL under inference delays, ensuring real-time learning and control (see \cref{figure:main_v2}).
    \item We propose \rthcp, a \textbf{R}eal-\textbf{T}ime \textbf{H}ybrid \textbf{C}ontrol with \textbf{P}hysics-informed model, a novel approach that combines MBRL, MFRL, and prior dynamics knowledge to enhance real-world robotic learning (see \cref{figure:rthcp}).
\end{enumerate}

To experimentally validate our methodology, we use the \furuta \cite{Furuta1992SwingupCO}, a rotary inverted pendulum that is a well-known benchmark in control theory due to its inherent instability and high-frequency control requirements. This feature makes it particularly challenging to learn to control in our context. We empirically demonstrate that \rthcp outperforms other model-based and model-free RL approaches in learning to control the robot within a limited amount of training data while satisfying real-time constraints.

\section{Related Work}

The deployment of RL in real-world applications introduces significant challenges \cite{dulac2019challenges}, particularly with respect to the inference time and the resulting delays in decision-making processes. 

{\textbf{Hybrid controllers:}} 
A recent line of research has emerged that focuses on improving the model-based planning process by integrating a policy \cite{
wang2019exploring}, a Q-function \cite{bhardwaj2020blending}, or both. \tdmpc \cite{Hansen2022tdmpc} combines both using a learned policy and Q-function alongside a data-driven model, while \phihp \cite{asri2024physics} applies a hybrid physics-informed model. 
Though these approaches help reduce inference delays in model-based methods, they do not eliminate them: a large inference time remains, particularly in the context of real robotics where the embedded computational resources are limited. Moreover, they have only been validated in simulation. In contrast, our method is designed to effectively address inference delays on a real robot with strong computational constraints.

{\textbf{Augmenting state methods:}}
Another line of research focuses on addressing delays within the MDP framework. Early work by Katsikopoulos et al. classified delays into three categories based on their occurrence: observation, action, and reward delays \cite{Katsikopoulos2003}. They also introduced the concept of delayed MDPs, where state augmentation helps restore the Markov property. Several subsequent studies have built upon this idea, applying state augmentation techniques to improve RL performance under delays \cite{Chen2020DelayAwareMR, Ramstedt2020ReinforcementLW}.
These methods typically augment the state with delayed actions to estimate the current state, effectively addressing transmission and execution delays, where state transitions of the environment remain synchronized with the agent’s action frequency. However, this approach is ineffective for inference delays, where the agent requires multiple time steps to compute an action, leading to a decision-making frequency that differs from the environment's natural frequency. In this work, we tackle this challenge by augmenting not only the state space but also the action space, enabling the agent to better compensate for inference delays.

{\textbf{Addressing inference delays:}} 
A third line of works specifically tackles inference delays. Ramstedt et al. \cite{Ramstedt2019RealTimeRL} introduce a framework to handle inference delays in real-world control tasks, while Xiao et al.  \cite{Xiao2020ThinkingWM} propose a continuous-time formulation of the Bellman equations to account for inference delays. However, these methods address inference delays that are shorter than a single time step, making them unsuitable for high-frequency control tasks where inference time spans multiple steps. In contrast, our work addresses a fundamentally different challenge: large inference delays, where the agent requires multiple control steps to compute an action.

{\textbf{Reinforcement Learning on Real Robots with Inference Delays:}}  
More closely related to our work, several studies have explored RL directly on real robots while addressing inference delays. A common approach consists in repeating the last executed action until a new action is computed, which can lead to degraded performance in dynamic environments.
To maintain a fixed action frequency despite the high inference times, an asynchronous RL approach has been proposed, decoupling policy updates from action execution \cite{Yuan2022AsynchronousRL}. More recently, staggered inference scheduling has been introduced, where multiple inferences are performed asynchronously \cite{Riemer2024EnablingRR}. However, the number of parallel inferences scales with inference time and is particularly constrained in robotic applications because of the limited resources of embedded computing hardware. In contrast, our work adopts a single-inference strategy designed to operate efficiently with limited computational resources.

\section{Background}
In this section, we briefly introduce the key concepts relevant to our approach: Markov Decision Processes (MDPs), Model Predictive Control (MPC) and the Cross-Entropy Method (CEM).

{\bf Markov Decision Processes:} 
Agent-environment interactions are often framed into the Markov Decision Process (MDP) framework, defined by the tuple \((\mathcal{S}, \mathcal{A}, \mathcal{T}, R, \gamma)\), where $\mathcal{S}$ is the state space, $\mathcal{A}$ the action space, $\mathcal{T}: \mathcal{S}\times \mathcal{A} \rightarrow \Pi(\mathcal{S})$ the transition function with $\Pi(\mathcal{S})$ the set of probability distributions over $\mathcal{S}$, $\mathcal{R}: \mathcal{S}\times \mathcal{A} \rightarrow \mathbb{R}$ the reward function, $\gamma \in [0,1]$ is a discount factor.
At each timestep $t$ the agent observes a state $s_t \in \mathcal{S}$ and picks an action $a_t \in \mathcal{A}$. Then, the environment returns a next state $s_{t+1} \sim \mathcal{T}(. | s_t, a_t)$ and a scalar reward $r_{t} = \mathcal{R}(s_t, a_t)$. The Markov property assumes that the transition $\mathcal{T}(. | s_t, a_t)$ depends only on the current state $s_t$ and action $a_t$, and not on prior states or actions.

{\bf Reinforcement Learning (RL):}
The objective in RL is to maximize the expected return $\sum_{t=t_0}^{\infty} \gamma^{t-t_0} r_t $ at each time step $t_0$. In model-free RL, an agent directly learns a policy, \ie a mapping $\pi : \mathcal{S} \rightarrow \mathcal{A}$ that maximizes this expected return. In contrast, in model-based RL, the agent learns a model $\hat{\mathcal{T}_\theta}$ of the transition function $\mathcal{T}$, then often uses this learned model $\hat{\mathcal{T}_\theta}$ to predict $s_{t+1}$ as $\hat{s}_{t+1} = \hat{\mathcal{T}_\theta}(s_t,a_t)$. The agent maximizes the expected return by optimizing a trajectory.

{\bf Model Predictive Control and Cross-Entropy Method:}
Model Predictive Control (MPC) is a control strategy that uses a model to predict next states and optimize a sequence of future control actions. MPC determines the best sequence of actions over a finite time horizon to mitigate a compounding error effect over time steps. The sequence of actions is often optimized using the Cross-Entropy Method (CEM), a gradient-free stochastic optimization algorithm commonly used in RL, particularly for complex and non-linear dynamics and reward functions. In the context of MPC, CEM iteratively samples action sequences, evaluates their performance, and updates a distribution over the action space by focusing on the best performing samples. This iterative process is repeated until convergence or until a maximum number of iterations is reached. 
Besides, MPC typically only executes the first action in the sequence before planning again, but it can also execute $d$ actions, in which case we call it {\em d-step MPC}. Leveraging d-step MPC is a key ingredient in the way we deal with inference delays.

The efficiency of CEM is characterized by two key parameters: the number of iterations \textbf{I} and the population size \textbf{P}, that directly impact the trade-off between inference time and performance. In practice, larger values improve performance but increase inference time. Therefore, introducing a method to optimize this trade-off is crucial to achieve real-time performance in real-world applications.

\section{Methods}

Inference delays occur when the time needed to determine a control command (called "action" below) is longer than the period at which the system requires it. 
To deal with inference delays, our method plans to obtain a sequence of actions long enough to cover the inference delay and stores these actions in a buffer from which the system can play them. We describe this approach more formally below, then we introduce \rthcp, a real-time hybrid planning method that integrates a physics-informed model with d-step MPC and a learned actor-critic agent, and we show that \rthcp adequately deals with inference delays.

\subsection{Planning to avoid execution gaps}

\textbf{Problem statement:} Regarding time, a plan-based control problem is characterized by the sampling period $\Delta_t$ at which the system needs to receive actions, a planning horizon $H^p$, and an inference time $T_i^{H^p}$ that depends on the planning horizon.
Inference delay occurs when computing an action takes longer than the sampling period, \ie $T_i^1 > \Delta_t$; otherwise, there is no inference delay. Besides, we assume that the inference time grows less than linearly with the planning horizon, meaning that there exists a horizon $H$ such that $T_i^H < H.\Delta_t$; otherwise, real-time control becomes infeasible. So, to make real-time control feasible, the planning horizon must be long enough, but not too long, as it affects inference time, which must be kept short to maintain real-time control.

\textbf{Impact of Inference Delays:} Inference delays cause actions to be executed in states different from those in which they were originally computed. Specifically, if an agent observes \(s_t\) but computes action \(a_t\) over \(d\) time steps, \(a_t\) is only executed when the environment has already transitioned to state \(s_{t+d}\). In this delayed setting, when the agent computes an action for a given state \(s_{t}\), it misses the intermediate states \([s_{t+1}, \ldots, s_{t+d-1}]\) over the next $d-1$ time steps. As a result, if state transitions occur at intervals of \(\Delta t\), the agent's observations and actions are effectively spaced by \(d \cdot \Delta t\). This leads to partial observability of the original MDP, as the agent loses information about the intermediate states between its consecutive observations.

\textbf{Our approach:} To address inference delays and restore the Markov property, we use a \textbf{d-step MPC} approach, which allows the agent to select a sequence of actions rather than a single action at each decision step. Then the system can continue executing precomputed actions from a buffer while new ones are being inferred, thereby avoiding execution gaps.
Additionally, we augment the observation space by incorporating the sequence of missed states along with the last sequence of actions that are not yet executed, enabling the agent to better track system dynamics despite the inference delay, as depicted in \cref{figure:dMPC}.

To formalize this process, considering an inference delay of \(d\) time steps, we define an augmented state representation \( s'_t = \{{{\textcolor{ForestGreen}{s_{d \cdot t}}}}, {{\textcolor{red}{s_{d \cdot t -(d-1)}, \ldots, s_{d \cdot t - 1}}}}, {{\textcolor{blue}{a_{d \cdot t}, \ldots, a_{d \cdot t + (d - 1)}}}}  \}\) which consists of the current state of the environment, the \(d-1\) missed states during inference, and the last sequence of actions not yet executed in the action buffer. At each decision step, the agent uses the history of these actions to estimate the future state \( {{\textcolor{ForestGreen}{s_{d \cdot t + d}}}}\), then it computes the optimal sequence of actions \(a'_{t+1}=[a_{d \cdot (t+1)}, \ldots, a_{d \cdot (t + 1) + (d-1)}]\) (see \cref{figure:dMPC}).

Figure \ref{figure:dMDP} illustrates how these augmented states are concatenated to restore the Markov property in the presence of inference delays. This ensures that the agent maintains a complete view of the system dynamics despite delayed decision-making.

To generalize this approach, we introduce a \dmdp \((\mathcal{S'}, \mathcal{A'}, \mathcal{T'}, \gamma, d)\) from the original MDP. This \dmdp augments the action space \(\mathcal{A'} = {\mathcal{A}}^d\) to plan multiple actions ahead, and augments the state space \(\mathcal{S'} = \mathcal{S}^d \cdot \mathcal{A}^d\) to restore the Markov property. The reward function is also augmented to \((\mathcal{R'} = {\mathcal{R}^d}\)), and the transition function is updated to \(\mathcal{T} = \mathcal{T}^d\). 

Building on this \dmdp, we introduce a framework for MBRL under inference delays, ensuring real-time learning and control.
Our approach unfolds as follows.

\par\smallskip $\bullet$ First, we determine on the system adequate values for $\Delta_t$ and $H^p$ such that $T_i^{H^p} < {H^p}.\Delta_t$. Depending on the context, $\Delta_t$ can be imposed by the system or tuned. Besides, $H^p$ must be tuned in all cases to ensure real-time execution and system stability.

\par\smallskip $\bullet$ Once $H^p$ is set, we measure $T_i^{H^p}$ on the system to characterize computational constraints.

\par\smallskip $\bullet$ Next, we set the execution horizon ${H^e}$ with ${H^e}_{min} \leq {H^e} \leq {H^p}$ and ${H^e}_{min} = \text{int}\left( \frac{T_i^{H^p}}{\Delta_t} \right) + 1$. ${H^e}_{min}$ is the smallest number of actions needed to feed the system during inference time ${T_i^{H^p}}$.

\par\smallskip $\bullet$ Finally, we apply d-step MPC with $d=H^e$. This strategy ensures that a full enough action buffer is always available while the system computes the next plan.

\subsection{Real-Time Robot Control with Physics-Informed Models}  

Planning with imperfect models generally leads to compounding errors, which degrade long-term decision-making. While MPC helps mitigate these errors through frequent replanning, it can introduce significant inference delays, especially when requiring long planning horizons. Using d-step MPC is a two-sided sword: from one side, it decreases the replanning frequency, which helps deal with inference delays. From the other side, it relies on longer open-loop trajectories, which may accumulate prediction errors if the model is inaccurate. Therefore, it is advantageous to i) reduce the execution horizon of d-step MPC, and ii) use a robust model able to accurately predict these transitions before replanning.

\begin{figure}[t]
\centering
\includegraphics[width=0.8\linewidth]{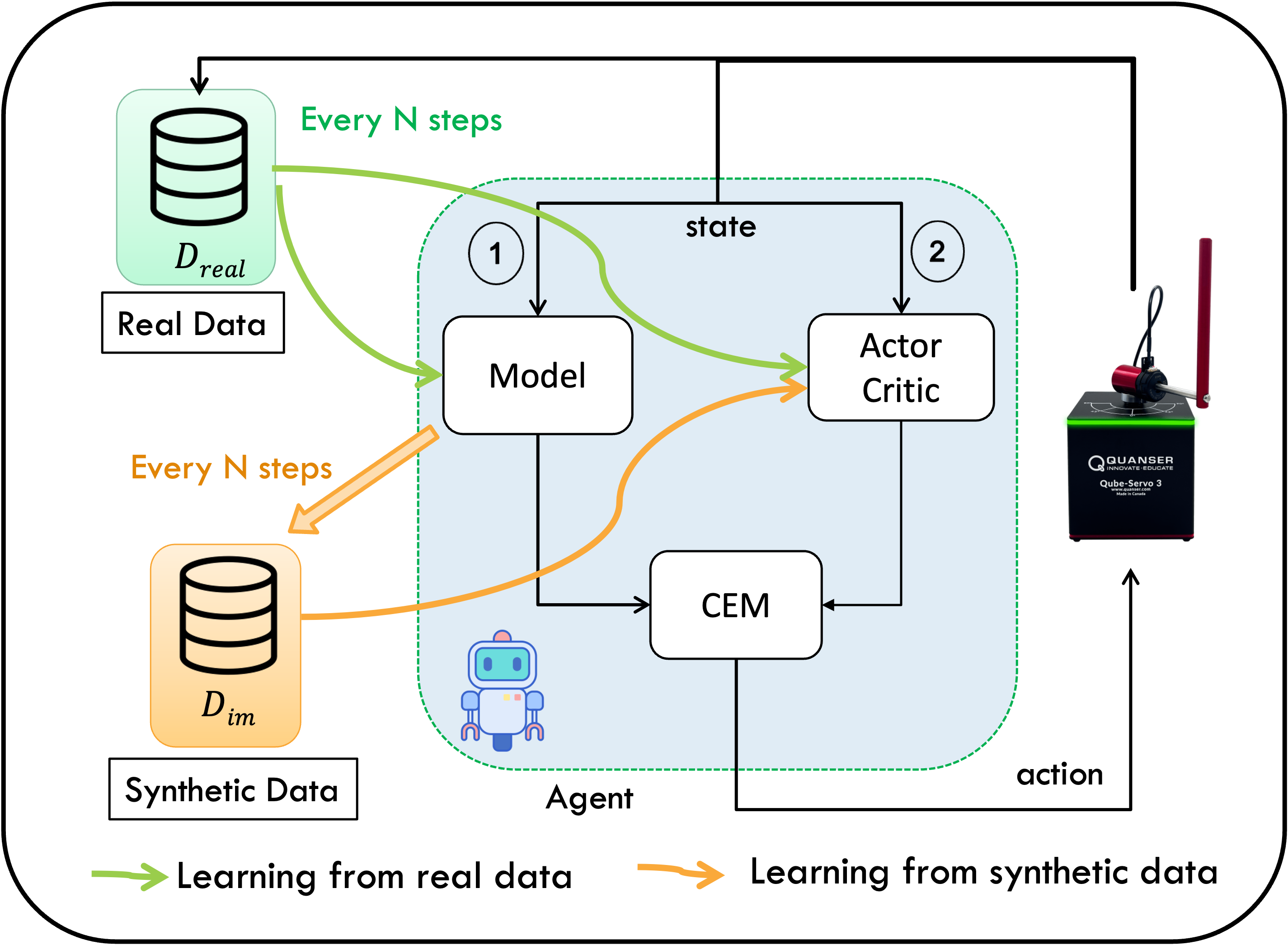}
\caption{Schematic description of \rthcp}
\label{figure:rthcp}
\end{figure}

To address these challenges, we introduce \rthcp, a \textbf{R}eal-\textbf{T}ime \textbf{H}ybrid \textbf{C}ontrol framework that integrates a \textbf{P}hysics-informed model, d-step MPC, and a learned actor-critic policy. \rthcp is designed to balance shorter execution horizons with robust trajectory predictions, ensuring both real-time execution and stability.
\rthcp follows an iterative real-time learning loop (see \cref{figure:rthcp}):

1) Online planning and data collection: The agent selects actions using hybrid planning, combining d-step MPC with an actor-critic policy $\pi$, interacts with the real system and stores transitions in \(D_{\text{real}}\).

2) Offline model and policy updates: Every \(N\) steps, offline updates of the physics-informed model and policy $\pi$ are performed using real-world transitions from \(D_{\text{real}}\).

3) Learning in imagination: The updated model periodically generates synthetic transitions into \(D_{\text{im}}\) to further refine $\pi$.

This real-time learning loop ensures that $\pi$ is continuously refined using both real-world and model-based data, achieving higher sample efficiency while maintaining real-time execution constraints. \rthcp is particularly well suited for real-time applications for two main reasons:

\textbf{(i) Reduced Inference Time through Hybrid Planning.}
The optimization problem in \rthcp is expressed as:
    \begin{equation}
    \begin{aligned}
    \label{eq:optimization}
    &A = \mathrm{arg} \underset{a_{t_0:t_0+H}}{\max}   \bigl(~\sum_{t=t_0}^{H} \gamma^{t-t_0} R({s}_t, a_t) +   \gamma^{H-t_0} Q(s_{t_0+H})~\bigr),\\ &\mathrm{subject~to} ~~~ {s}_{t+1} = \hat{\mathcal{T}_\theta}(s_t, a_t).
    \end{aligned}
    \end{equation}
\rthcp accelerates the convergence of CEM by using informative action candidates from $\pi$, reducing both the population size and the number of iterations. In addition, it allows shorter planning horizons \(H^p\) by using immediate reward up to a reduced horizon \(H^p\), and incorporating a Q function to estimate the long-term return. 
Since the delay \(d\) depends on the inference time, which itself strongly depends on both the planning horizon \(H^p\) and the population size in CEM, this hybrid planning strategy ultimately reduces delays.

\textbf{(ii) Mitigating compounding errors with a robust physics-informed model.} Unlike purely data-driven models that struggle with trajectory predictions outside their training distribution, particularly for long horizons, \rthcp uses a physics-based prior that remains valid across the entire state and action spaces. This allows \rthcp to consistently predict accurate trajectories, thereby reducing prediction errors for the hidden states in the \dmdp (see \cref{figure:predict}).

Our approach builds upon \phihp \cite{asri2024physics}, which leverages a physics prior in a hybrid planning scheme. 
However, \rthcp introduces key modifications to \phihp, making it suitable for real-time applications, particularly in how trajectories are collected and used for learning.
Specifically, \phihp relies solely on MPC for real data collection, requiring long planning horizons that increase inference time and introduce delays. In contrast, \rthcp uses a hybrid planning approach to reduce delays. In addition, \phihp trains $\pi$ exclusively on model-generated rollouts without use of the available real data. In contrast, \rthcp jointly trains the model and $\pi$ on real data while iteratively refining $\pi$ using imagined rollouts.

\section{Experimental Study}

\subsection{Experimental setup}
{\bf Environment and robotic platform:}
In this work, we apply our method to a real \furuta, an under-actuated system, commonly used to evaluate control strategies for nonlinear and unstable dynamics. The \furuta consists of a rotary arm actuated by a motor and a pendulum attached to the end of the arm. The goal is to swing up the pendulum from its downward position and stabilize it upright while respecting mechanical constraints.  

We define the system state at each time step \(t\) as  
$
s_t = (\alpha_t, \beta_t, \dot{\alpha}_t, \dot{\beta}_t)
$, 
where \(\alpha_t\) is the rotary arm angle, \(\beta_t\) is the {pendulum angle, \(\dot{\alpha}_t\) and \(\dot{\beta}_t\) are their angular velocities. The control action is the applied voltage to the motor, which determines the generated torque $a_t = V_t$.

Our experiments allow us to evaluate the proposed method under real-world physical conditions, including stochastic dynamics, computational constraints, and inference delays.

\begin{figure}[h]
\centering
\includegraphics[width=0.9\linewidth]{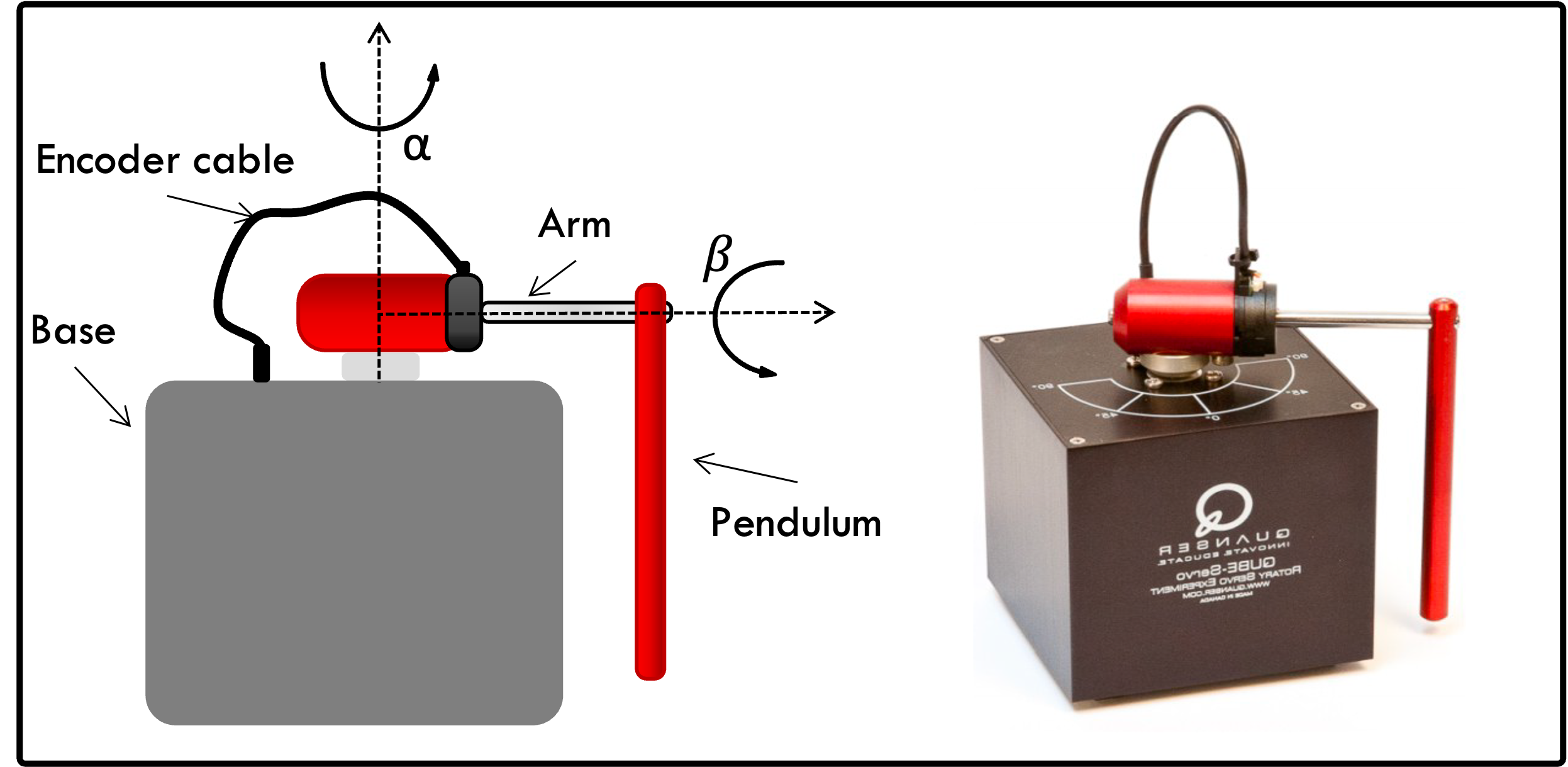}
\caption{The Furuta pendulum (schematic on the left, experimental setup on the right)}
\label{fig:pendulum}
\end{figure}

{\bf RT-HCP implementation:}
To model the dynamics of the Furuta pendulum, we consider a frictionless two-link rotary inverted pendulum (RIP) model. We describe the physical prior using the following equations of motion derived from the Euler-Lagrange formulation:
$$\begin{pmatrix}\ddot{\alpha}\\ \ddot{\beta}\end{pmatrix}=\begin{pmatrix}
   
    M^{-1}(\beta)\left(\begin{pmatrix}\tau(a)   \\ 0 \end{pmatrix}-N(\beta, \dot{\alpha}, \dot{\beta}) - G(\beta)\right)
\end{pmatrix}$$ where:
$$M(\beta)=\begin{pmatrix}J_1+\frac{1}{4}m_pL_p^2\sin^2(\beta) & \frac{1}{2}m_pL_rL_p\cos(\beta)\\\frac{1}{2}m_pL_rL_p\cos(\beta) & J_2\end{pmatrix},$$
$$N(\beta,\dot{\alpha},\dot{\beta})= \frac{1}{2}m_pL_p\sin(\beta)\begin{pmatrix}L_p\dot{\alpha}\dot{\beta}\cos(\beta)-L_r\dot{\beta}^2\\-\frac{1}{2}L_p\dot{\alpha}^2\cos(\beta)\end{pmatrix}, $$
$$G(\beta)=\begin{pmatrix}0\\\frac{1}{2}m_pgL_p\sin(\beta)\end{pmatrix},
  \tau(a)=\dfrac{k_t(-a-k_m\dot{\alpha})}{R_m}.$$

The parameters $m_p$, $L_p$, and $J_p$ are the mass, length, and moment of inertia of the pendulum respectively. The same parameters for the rotor are $m_r$, $L_r$, and $J_r$. $k_t$, $ k_m$, $R_m$, and $u$ are the motor torque constant, electromotive force constant, electric resistance, and input voltage respectively. 

Although this analytical model provides a first-principles approximation of the \furuta behavior, it does not account for some real-world effects that significantly impact the dynamics, such as: 1) joint friction and actuator delays, which introduce resistance and latency in motion;  2) encoder cable effects, where the physical cable connecting the encoder to the base introduces additional damping forces and tension, particularly during large oscillations. This results in nonlinear perturbations that alter the expected motion of the system.  
Furthermore, the parameters of this system are only approximately known, which introduces some errors. To mitigate these errors, the \rthcp model captures a significant part of the dynamics with the prior model above, while compensating for unmodeled effects through a residual neural network. This network is implemented as a 4-layer MLP with 16 neurons. For planning, we use CEM with $\textbf{I} = 3$ and $\textbf{P} = 500$, of which 50 from $\pi$.

{\bf Baselines:}
We compare \rthcp against three state-of-the-art RL methods. For a fair comparison, we adapt all model-based baselines using our real-time inference delay management approach:

$\bullet$ \textbf{TD3} \cite{fujimoto2018addressing}, a model-free RL algorithm known for its strong performance in continuous control tasks. \tddd is fast enough to avoid inference delays.

$\bullet$ \textbf{RT-TDMPC} \cite{Hansen2022tdmpc}, a state-of-the-art hybrid MBRL/MFRL algorithm that performs local trajectory optimization (planning) with a learned world model and a policy. \tdmpc 
relies on a purely data-driven model, whereas \rthcp uses a physics-informed model.
We use the original hyper-parameters and adapt the original implementation to support early termination in the \furuta environment.

$\bullet$ \textbf{RT-PETS} \cite{chua2018deep}, a well-established model-based RL algorithm that employs probabilistic ensemble models to generate trajectory samples. \pets is known for its good sample efficiency and robustness to model uncertainty. We use the pytorch implementation from \cite{pets2019imp} with the original hyper-parameters.
\par\noindent
\subsection{Experimental Results}

In this section, we empirically measure the inference time of all methods on the real robot to determine the inference delay. Then we train these methods on the \furuta with a fixed budget of 200k training steps. 
\par\noindent
\textbf{Extracting Real-Time Constraints:}
To assess the impact of inference delay, 
We report in \cref{table:inference} the empirical inference times \(T_i\) measured on the real robot for each method, as a function of the planning horizon \(H^p\), with the sampling period \(\Delta t = 20\) ms. \(T_i\) includes both action selection and online updates when present (\eg for \tddd), but excludes offline updates.
We also compute the corresponding inference delay (\(d = T_i / \Delta t\)) and determine the minimum execution horizon \({H^e}_{\min}\), which ensures that a precomputed action sequence covers the entire delay. Being model-free, \tddd has the lowest inference time. In contrast, being model-based \rttdmpc, \rtpets and \rthcp require longer inference times due to their trajectory optimization component. 

\begin{table}[ht]
\caption{Empirical measurements of inference time $T_i$, inference delay $d$ and execution horizon $H^e$ as a function of the planning horizon $H^p$ for $\Delta t = 20$ ms.}
\label{table:inference}

\begin{center}
\resizebox{0.9\linewidth}{!}
{
\begin{tabular}[b]{|l|c|c|c|c|}
\hline
Method & \tddd  & \rttdmpc & \rtpets &  \rthcp \\
\hline

\({H^p}\) & 1  & 5 & 15 &  5\\
\hline

\(T_i (ms)\) & \textbf{ 16 $\pm1$} & 47 $\pm2$ & 156 $\pm2$ & 36 $\pm4$\\
\hline

delay & \textbf{0,8}  & 2,35 & 7,81 & 1,8 \\
\hline

\({H^e}_{min}\) & 1  & 3 & 8 & 2\\
\hline 

\end{tabular}
}
\end{center}
\end{table}

\par\noindent
\textbf{Learning Performance Under Real-Time Constraints:} 
We evaluate the learning performance of each method when deployed directly on the robot under real-time constraints. \cref{figure:learning} presents the evolution of episodic reward throughout training. The results are averaged over 10 evaluation episodes using a single policy trained with a fixed random seed. The results show that \rthcp achieves higher sample efficiency and better overall performance than \tdmpc and \tddd. Specifically, \rthcp reaches a threshold reward of 300, which is sufficient to successfully complete the task, after only 60k training steps (20 minutes). In contrast, \tdmpc and \tddd require respectively 100k (33 mins) and 160k (53 mins) to reach the same performance level. Furthermore, \rthcp consistently outperforms \tddd and \rtpets throughout training while also outperforming \tdmpc in the early stages and remaining competitive towards the end. This highlights its effectiveness in real-world training conditions, making it a robust choice for real-time robotic learning.

\begin{figure}[ht]
\centering
\includegraphics[width=\linewidth]{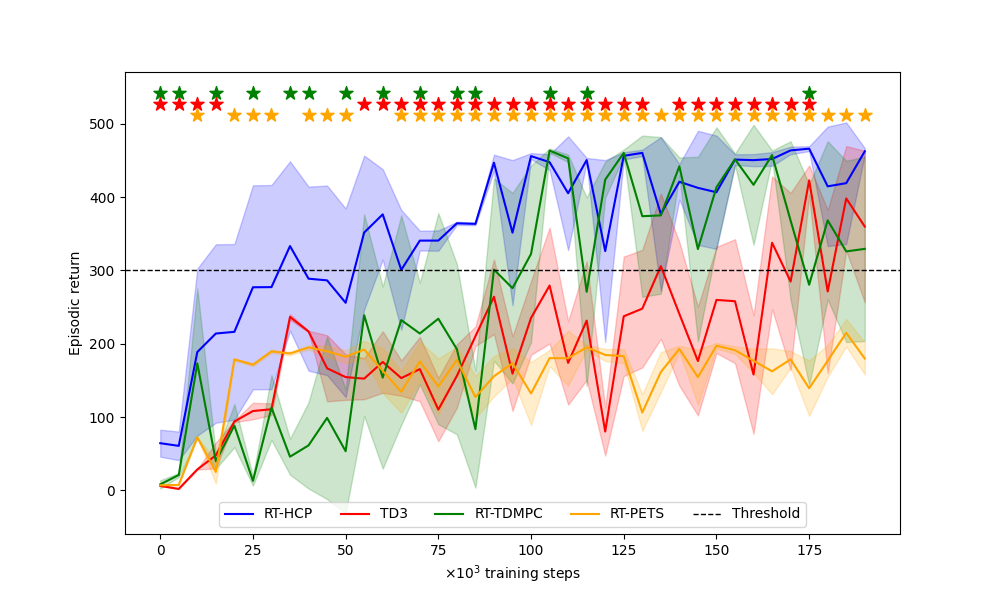}
\caption{Learning curves on the real \furuta under real-time constraints. Mean and 95\% confidence intervals over 10 evaluations.
Colored stars indicate statistically  significant differences using Welch’s t-test between \rthcp (blue) and each baseline, with the same color as the baseline.}
\label{figure:learning}
\end{figure}
\par\noindent
\textbf{Control Performance and Stabilization}  
We compare the control performance of each method at the end of training.
\cref{figure:traj} and \cref{table:time_swingup} show that \rthcp  achieves the fastest and most stable swing-up, reaching the upright position ($\alpha=\pm\pi$) earlier while staying closer to the center ($\beta=0$).  In contrast, \tdmpc and \tddd require more time and exhibit more oscillations, deviating further from the center before stabilization. Furthermore, \pets fails to complete the task, never reaching the upright position.

\begin{table}[ht]
\caption{Average time required to reach the upright position and the rotor deviation from the center after swing-up. Mean and std. over 10 trajectories. \rtpets never succeeds.}
\begin{center}
\centering
\resizebox{\linewidth}{!}{
\begin{tabular}{|l|c|c|c|c|}
    \hline
    Method & \tddd & \rttdmpc & \rtpets  & \rthcp    \\
    \hline
    Swing-up time (s) & 2.45
 $\pm 0.51
$ & 2.28
 $\pm0.80
$ & $+\infty$ & \textbf{1.67
 $\pm0.97
$}\\ \hline

Rotor deviation (rad) & 1.06
 $\pm 0.25
$ & 0.83
 $\pm0.32
$ & $+\infty$ & \textbf{0.53
 $\pm0.28
$}\\
\hline
\end{tabular}
}    
\end{center}
\label{table:time_swingup}
\end{table}

\begin{figure}[ht]
\centering
\includegraphics[width=0.9\linewidth]{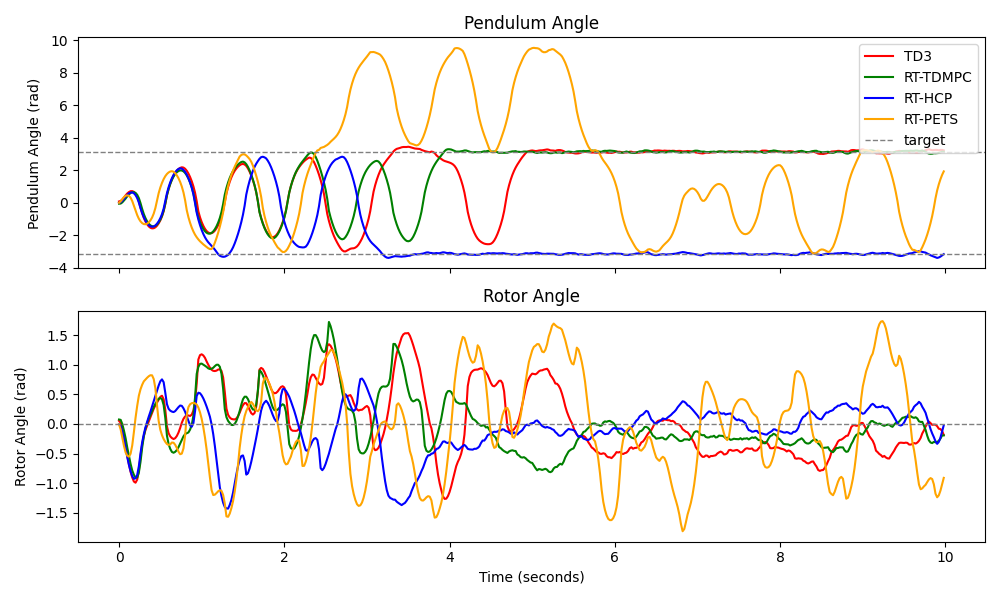}
\caption{Representative trajectory for each agent, showing the evolution of the pendulum  (a) and rotor (b) angles over time. Refer to the video for visualizations of the learned policies.}
\label{figure:traj}
\end{figure}

\par\noindent
{\bf Impact of delay on performance:}
To examine how delays affect performance, we compare the performance variation across different planning horizons using the d-step MPC framework.
\cref{figure:horizon} illustrates that \rthcp degrades less with increasing planning horizon compared to \rttdmpc. This difference can be attributed to two key factors. First, \rthcp benefits from smaller neural networks and a smaller CEM population compared to \rttdmpc, leading to faster inference. Second, the physics-informed model of \rthcp allows it to predict future states more accurately, mitigating the accumulation of prediction errors over time. Conversely, \rttdmpc relies solely on data-driven models, which are more susceptible to compounding errors over time. Contrary to \tdmpc and \rthcp, \rtpets is less effective with short horizons, as it does not rely on a Q-function, and thus requires longer horizons for good performance.

\begin{figure}[h]
\centering
\includegraphics[width=0.9\linewidth]{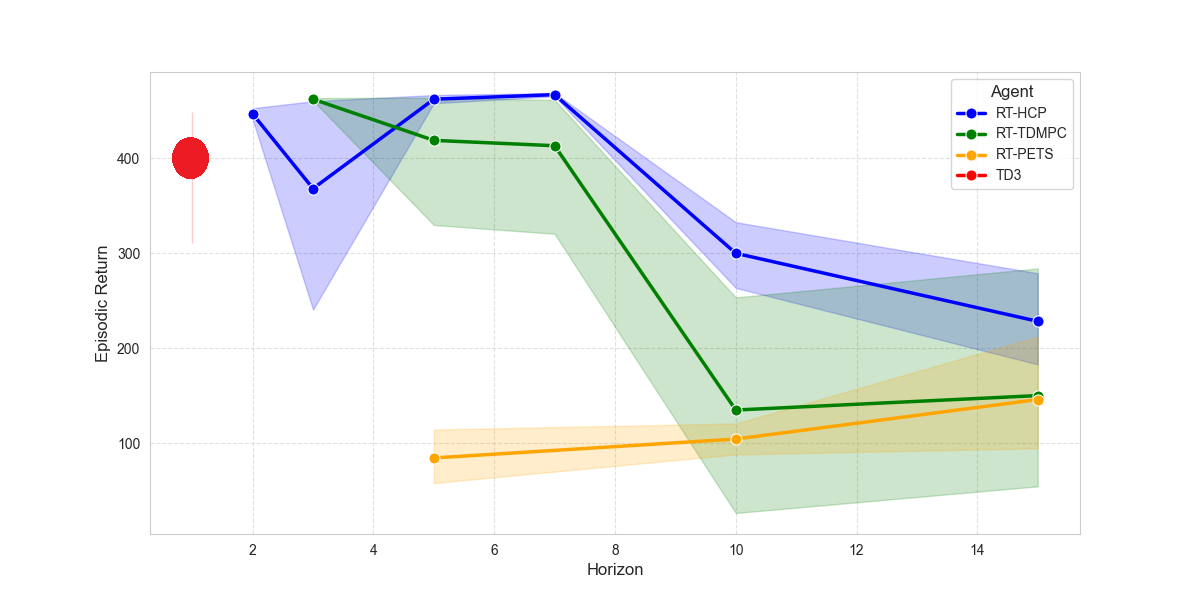}
\caption{Evolution of the average return as a function of the horizon. Mean and 95\% confidence interval over 10 trials.}
\label{figure:horizon}
\end{figure}

\par\noindent
\textbf{Model Prediction Accuracy:}  
To assess the accuracy of the learned models, we evaluate their ability to predict system trajectories given an initial state and a predefined sequence of actions. As shown in \cref{figure:predict}, \rthcp provides the most accurate trajectory predictions, closely matching the ground truth, demonstrating the benefit of the physics-informed model in reducing prediction errors. In contrast, \tdmpc exhibits the largest deviations over time, leading to cumulative errors that degrade long-term planning. Meanwhile, \pets achieves better trajectory predictions than \tdmpc, demonstrating the effectiveness of its ensemble-based approach in modeling system dynamics. However, despite its improved predictive accuracy, \pets fails to complete the swing-up task. This reinforces the idea that while accurate trajectory predictions are important, a large inference time can still prevent an agent from succeeding, even with a good predictive model.

\begin{figure}[ht]
\centering
\includegraphics[width=0.9\linewidth]{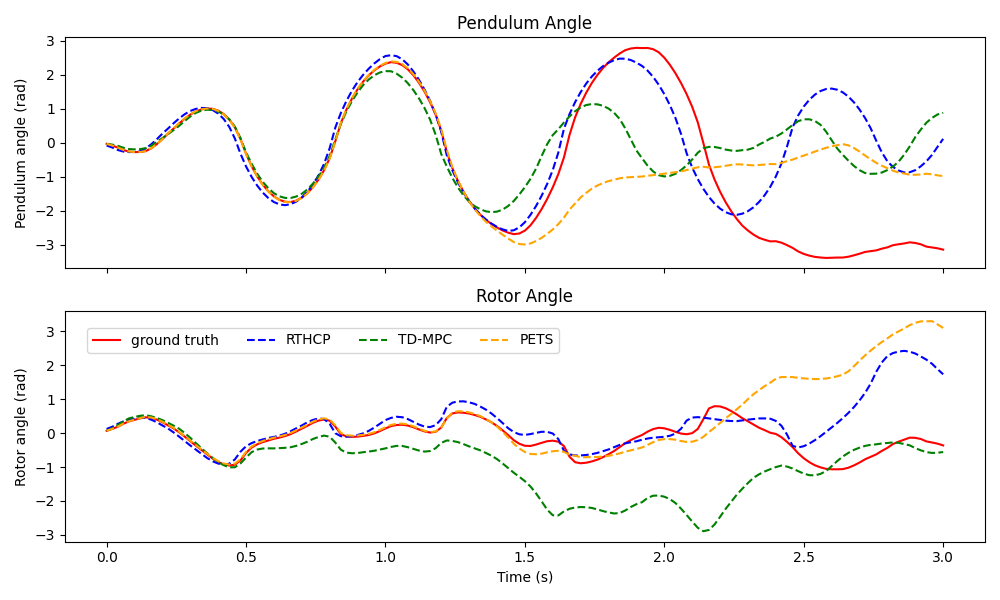}
\caption{Trajectory predictions vs. ground truth from the same initial state and actions. \rthcp better approximates the system dynamics, closely matching the ground truth, while \tdmpc exhibits larger deviations over time.}
\label{figure:predict}
\end{figure}

\section{Discussion and conclusion}
Using our framework to deal with inference delays, we could fairly compare the performance of model-free and model-based reinforcement learning approaches directly on a real robot.
From these comparisons, we have learned the following: (i) model-free methods like \tddd benefit from a short inference time but require a lot of samples; (ii) Model-based RL methods like \pets improve sample efficiency but at the cost of longer inference times, preventing high-frequency control and leading to failure despite accurate predictions, as inference delays remain a major challenge.
(iii) Hybrid methods such as \tdmpc partially reduce inference time but do not eliminate it, causing persistent delays and making them too sensitive to model inaccuracies for stable control.
In addition to these findings, we have presented \rthcp as an intermediate approach that shines in two aspects.

First, in terms of learning efficiency, the combination of physics-informed and data-driven model acquisition significantly improves the sample efficiency with respect to \pets and \tdmpc.
Second, the policy obtained at the end of the learning process performs better than those of \pets and \tdmpc because it allows it to provide actions at a higher frequency. Under a constrained learning budget, which is mandatory when learning on a robot, it also outperforms the one from \tddd, despite the shorter inference time of the latter.
Our study was limited to the context where the state of the robot was directly accessible for control. In the near future, we intend to extend our work to the case where the input of the controller is an image, leveraging recent progress in vision-based RL.

\section*{Acknowledgments}
The authors thank Pascal Morin for his help with the robot. This research was funded, in whole or in part, by the European Commission's Horizon Europe Framework Programme, under the PILLAR-robots project (grant agreement No 101070381), and by l’Agence Nationale de la Recherche (ANR) under the RODEO project (ANR-24-CE23-5886) and PEPR Sharp (ANR-23-PEIA-0008, FRANCE 2030). 

\bibliographystyle{IEEEtran}
\bibliography{IEEEabrv,references}

\end{document}